\documentclass[conference]{IEEEtran}
\IEEEoverridecommandlockouts

\usepackage{cite}
\usepackage{amsmath,amssymb,amsfonts}
\usepackage{algorithmic}
\usepackage{graphicx}
\usepackage{textcomp}
\usepackage{xcolor}
\usepackage{placeins}
\usepackage{times}
\usepackage[utf8]{vietnam}
\usepackage[utf8]{inputenc}
\usepackage[main=english, vietnamese]{babel}
\babeltags{vi=vietnamese}
\usepackage{type1cm}
\usepackage[T1]{fontenc}
\usepackage{microtype}
\usepackage{multirow}
\usepackage{graphicx}
\usepackage{marvosym}
\usepackage{float}
\usepackage{amsmath,amssymb,amsfonts}
\usepackage{pifont}
\usepackage{siunitx}
\usepackage{latexsym}
\usepackage{etoolbox}

\usepackage[bookmarks,bookmarksopen,bookmarksdepth=3,bookmarksnumbered]{hyperref}
\hypersetup{colorlinks, citecolor=blue, linkcolor=blue, urlcolor=blue}

\makeatletter
\newcommand{\linebreakand}{%
  \end{@IEEEauthorhalign}
  \hfill\mbox{}\par
  \mbox{}\hfill\begin{@IEEEauthorhalign}
}
\makeatother

\begin{document}

\title{Integrating Semantic Information into\\Sketchy Reading Module of Retro-Reader for\\Vietnamese Machine Reading Comprehension}

\author{\IEEEauthorblockN{\textbf{Hang Thi-Thu Le$^{\text{1, 2}}$, Viet-Duc Ho$^{\text{1, 2}}$, Duc-Vu Nguyen$^{\text{*, 1, 2}}$, Ngan Luu-Thuy Nguyen$^{\text{1, 2}}$}}
\IEEEauthorblockA{$^{\text{1}}$University of Information Technology, Ho Chi Minh City, Vietnam \\
$^{\text{2}}$Vietnam National University, Ho Chi Minh City, Vietnam\\
\texttt{\{18520274,18520610\}@gm.uit.edu.vn}\qquad \texttt{\{vund,ngannlt\}@uit.edu.vn}}
\thanks{$^{\text{*}}$Corresponding author}
}

\maketitle

\begin{abstract} 

Machine Reading Comprehension has become one of the most advanced and popular research topics in the fields of Natural Language Processing in recent years. The classification of answerability questions is a relatively significant sub-task in machine reading comprehension; however, there haven’t been many studies. Retro-Reader is one of the studies that has solved this problem effectively. However, the encoders of most traditional machine reading comprehension models in general and Retro-Reader, in particular, have not been able to exploit the contextual semantic information of the context completely. Inspired by SemBERT, we use semantic role labels from the Semantic Role Labeling (SRL) task to add semantics to pre-trained language models such as mBERT, XLM-R, PhoBERT. This experiment was conducted to compare the influence of semantics on the classification of answerability for the Vietnamese machine reading comprehension. Additionally, we hope this experiment will enhance the encoder for the Retro-Reader model's Sketchy Reading Module. The improved Retro-Reader model’s encoder with semantics was first applied to the Vietnamese Machine Reading Comprehension task and obtained positive results.

\end{abstract}


\section{Introduction}

Machine reading comprehension (MRC) is one of the challenging tasks of natural language processing; This task requires the model to answer questions based on specific passages \cite{hermann2015teaching, joshi2017triviaqa}. In order to give accurate answers to the questions in the MRC task, the model needs to be able to read and understand the model language at a high level. In recent years, the MRC’s model must not only deal well with giving correct answers \cite{lewis2018generative, lee2018ranking} to questions but also distinguish unanswerable questions to avoid giving inappropriate answers \cite{rajpurkar2018know}.

Determining the ability to answer in machine reading comprehension tasks is essential; earlier studies have partly solved this problem in English \cite{hu2019read+, back2019neurquri}. Typically, the Retro-Reader improvement model of Zhang et al. \cite{retro} has obtained state-of-the-art results compared to previous studies.  

According to \cite{retro}, so far, standard reading systems consist of two modules: 1) an encoder with a robust language model and 2) a decoder cleverly designed to match the task characteristics of the MRC. Retrospective Reader (Retro-Reader) has designed a reading comprehension model that enhances the classification problem by integrating two phases of reading and verification strategies. However, this model has been too focused on the decoder and not enough on the encoder aspect.

Pre-trained language models (PrLMs) such as ELMo \cite{peters-etal-2018-deep}, GPT \cite{radford2018improving}, BERT \cite{devlin2018bert}, RoBERTa \cite{liu2019roberta}, ALBERT  \cite{lan2019albert}, which have succeeded in various natural language processing tasks, are well-known and play the role of a powerful encoder. However, existing language models only focus on exploiting the plain contextual features, rarely considering explicit contextual semantic clues. That may be one of the shortcomings of current language representation models.

By integrating the plain context representation on BERT and explicit semantics from semantic role labels for deeper meaning representation, Zhang at el. \cite{zhang2020semantics} obtained state-of-the-art results compared to regular BERT. This has demonstrated that providing explicit semantics to pre-trained language models is necessary. 

Many studies in English utilize semantic information from semantic role labels to improve Natural Language Understanding (NLU) tasks. However, there is very little research on SRL tasks in Vietnamese, and those that exist mainly concentrate on developing semantic labeling systems
\cite{hoang2015building, pham2015building, le2018vietnamese} without using the semantics of semantic role labels for other tasks. That is the motivation for us to do this study.

Inspired by SemBERT \cite{zhang2020semantics} and motivated by the aim to improve the encoder for the Retro-Reader model, we propose to implement a Retro-Reader model that includes semantic information to compare the influence of semantics on the classification of answerability for Vietnamese MRC. So, we integrate semantic information into the encoder of the classifier module in Retro-Reader.

The primary contributions of our current research:
\begin{itemize}
    \item Comparing the influence of semantics from semantic role labeling task on classifying answerability of questions for Vietnamese machine reading comprehension.
    \item Improved encoder for Sketchy Reading Module of Retro-Reader model.
    \item Demonstrate the significance of semantic role labels in Vietnamese.
\end{itemize}

\section{Related Works}

\subsection{Machine Reading Comprehension}

The origins of the MRC extend back to the 1970s when researchers started to recognize the significance of text understanding in artificial intelligence. The first reading comprehension system proposed by W Lehnert in 1977, QUALM \cite{lehnert1977process}, established the foundation for developing later machine reading comprehension research.

Machine reading comprehension was clearly described and handled as a supervised learning task by 2013: given a passage and a relevant question, the model should give the answer in the required format. The MRC task has long been in development in English, with the early introduction of datasets \cite{rajpurkar2016squad, rajpurkar2018know}. In contrast, the MRC problem in Vietnamese has only begun to receive attention, with several datasets such as UIT-ViQuAD1.0 \cite{viquad1}, ViMMRC \cite{vimmrc}, UIT-ViNewsQA \cite{vinewsqa}. Along with that, many studies on Vietnamese MRC systems have been developed \cite{kadlec2016text, wang2017gated}. One problem encountered by the above MRC systems is that these MRC systems are designed with the assumption that all questions can be answered but this is not always the case in actual life. Consequently, the researchers added unanswerable questions to the MRC datasets later on \cite{rajpurkar2018know, trischler2016newsqa, viquad2}. This is a significant challenge for later MRC models; a successful MRC model must be able to properly handle two aspects: providing accurate answers to answerable questions and identifying the answerability of the question.

Distinguishing unanswerable questions is a relatively important task in Automatic Reading Comprehension, but previous studies have not paid attention to this issue or only used simple methods. Most of the question's answerability classification tasks are trained with answer span  prediction. Liu et al. \cite{liu2018stochastic} added a [CLS] token to the context, and used an additional layer of simple classifiers to their MRC model, Back et al. \cite{back2019neurquri} relied on the attention mechanism to identify the words in the question that made them impossible to answer, thereby calculating a score for the questions and determining answerability based on that score.

Unlike previous studies, the Retro-Reader  \cite{retro} model takes an entirely new approach with two strategic phases: read and verify. It obtained state-of-the-art results on the MRC in English. In this study, we will enhance the Retro-Reader model and implement it for the Vietnamese MRC.

\subsection{Explicit Contextual Semantics}

The semantic role labeling task aims to identify the predicate-argument structure of a sentence by analyzing the shallow semantics of natural language documents. It assists us in finding the answers to questions like "Who?" "Did what?" "For whom?" "With what?" "Where?" and "When?" \cite{li2019dependency,li2018unified}. Coincidentally, those are the same issues that MRC tasks in general or QA in particular have to deal with; it is entirely fair to apply semantic role labels for the MRC task.

In 2004, Narayanan and Harabagiu \cite{narayanan2004question} proposed employing semantic roles in question answering (QA) systems; this was one of the earliest research on using semantic role labels as a side task in QA. Most subsequent studies \cite{fliedner2007linguistically, ofoghi2006hybrid} that used semantic roles as a primary method in the QA process yielded numerous positive outcomes. As a result of the dominance of pre-trained language models  \cite{peters-etal-2018-deep, radford2018improving, devlin2018bert, liu2019roberta}, semantic information from semantic role labels is no longer valued or utilized. However, semantic labels are helpful in many other NLP applications \cite{mihaylov2016discourse, shi2016knowledge}.

In 2020, Zhang et al. \cite{zhang2020semantics} proposed incorporating explicit contextual semantics from the pre-trained semantic role labeling task and introducing a more effective language representation model called SemBERT. It achieves state-of-the-art and significantly improves results on ten reading comprehension and linguistic inference tasks in English. 

There has not been any research on Vietnamese using semantic information from semantic labels to enhance MRC/QA tasks or NLU tasks. To compare the impact of semantics on the classification of answerability for the Vietnamese MRC task, we will use SemBERT's method to create variations of SemBERT from several pre-trained language models.

\section{Methodology}

\subsection{Retro-Reader Model}

\begin{figure*}[!ht]
\centering
\includegraphics[scale=0.6]{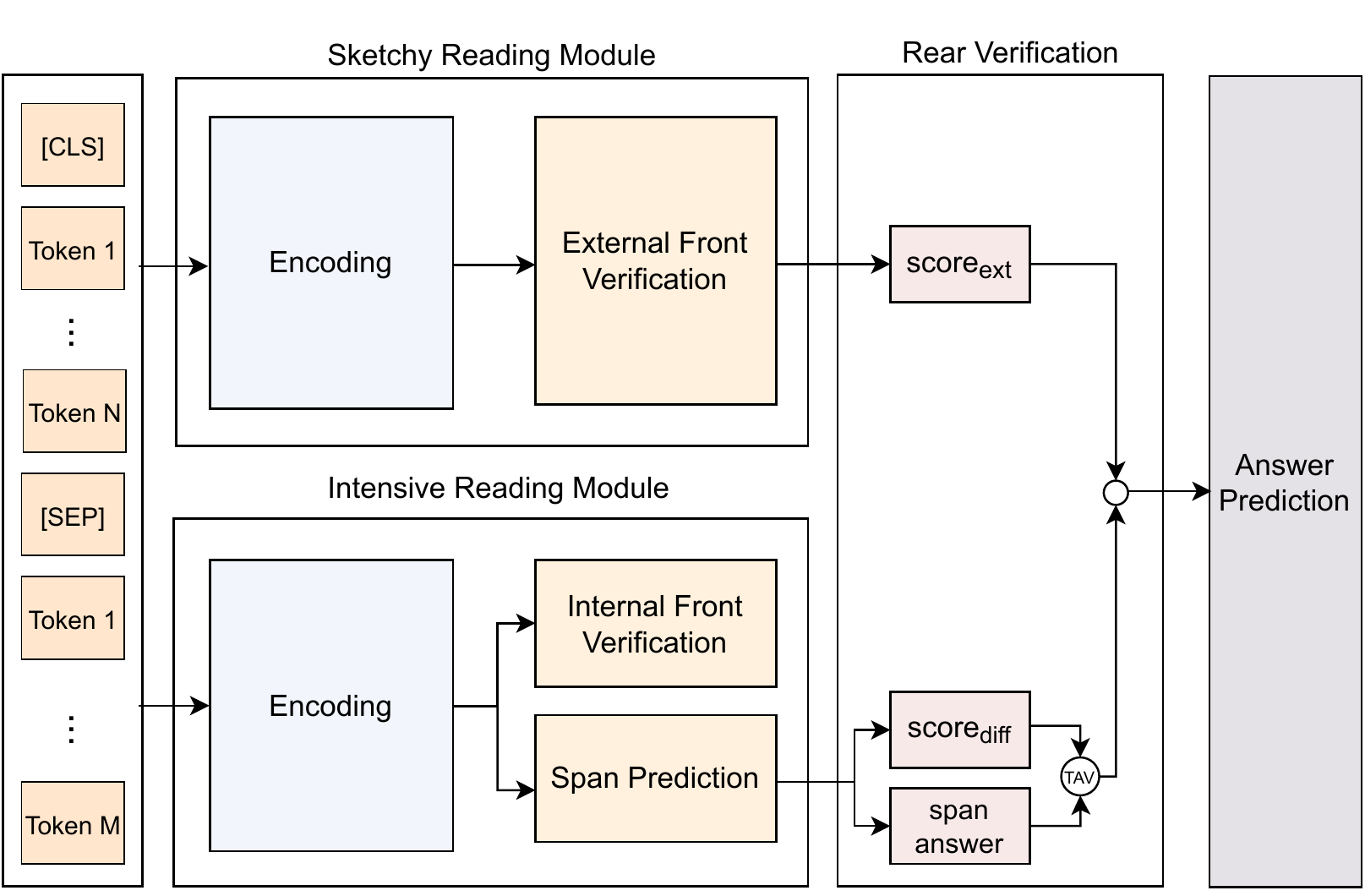}
\caption{Overview of the Retrospective Reader model \cite{retro}}
\label{fig:Retro}
\end{figure*}

The retrospective reader (Retro-Reader) \cite{retro} has two parallel modules that can be used to execute a two-stage reading process: a \textit{sketchy reading module} and an \textit{intensive reading module}. Then, combine the answerability confidence score (intensive reading module) with the judgment score (sketchy reading module) to yield the final answer called \textit{rear verification}.  In both modules, \cite{retro} apply BERT-based PrLMs as Encoder to acquire contextual representations of input tokens. An overview of the Retro-Reader model is shown in figure \ref{fig:Retro}.

\textbf{\textit{Sketchy reading module}} can be seen as a classifier to verify the answerability of the question; the final hidden state is passed through a fully connected layer to get classification logits composed of answerable and unanswerable elements. The output score differs between the no-answer score and the answer score.

\textbf{\textit{Intensive reading module}} with the aims of verifying answerability, predicting the answer span, and giving the final predicted answer. Use the final layer hidden state to predict each token's start and end probabilities, thereby giving the predicted answer span and no-answer score. In addition, there is an internal front verifier such that the intensive reader can also identify unanswerable questions.

\textbf{\textit{Rear verification}} combines the output of the two modules above, which is an aggregated verification for the final answer.

\subsection{Semantic-aware BERT}

\begin{figure*}[ht]
\centering
\includegraphics[scale=0.535]{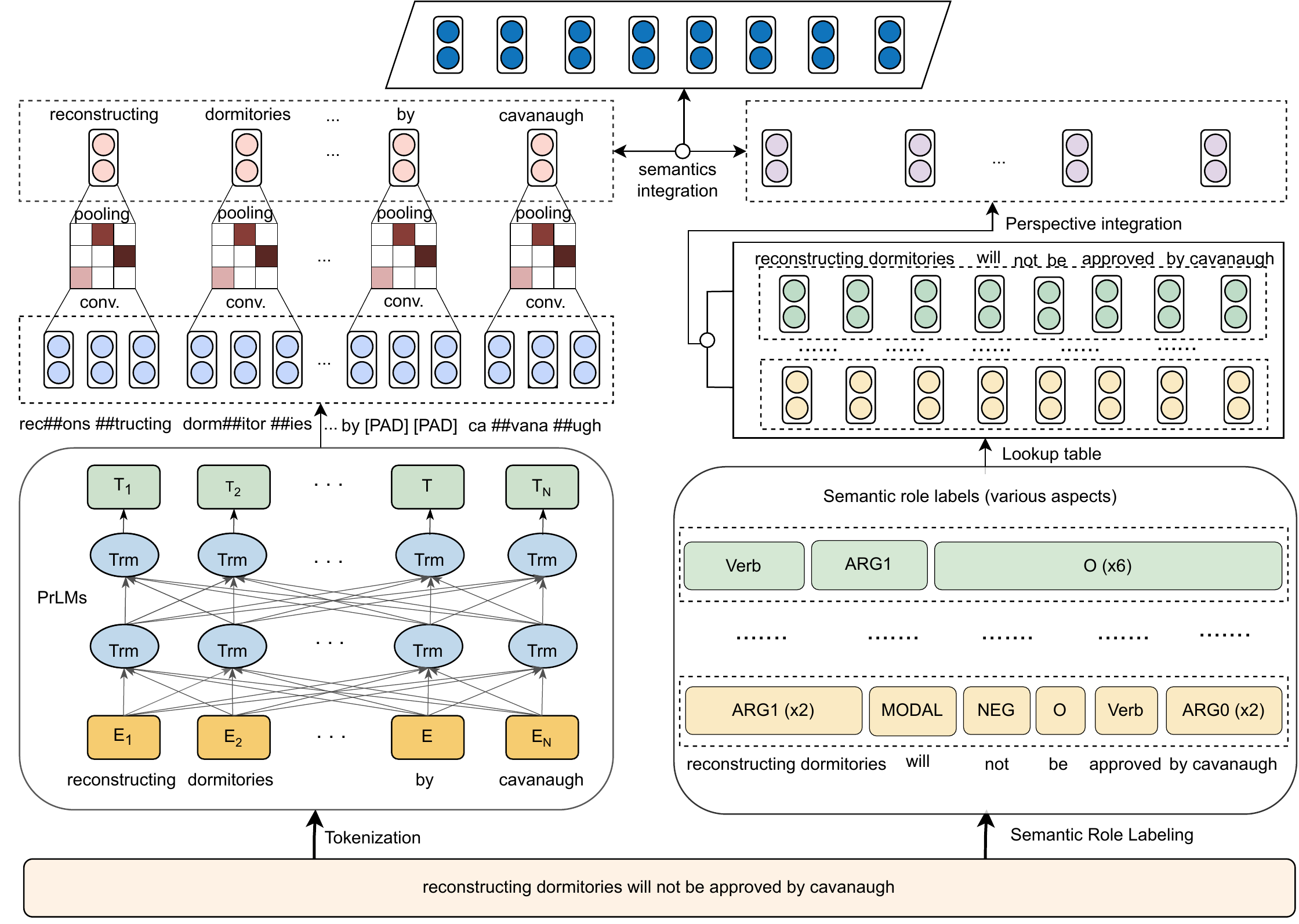}
\caption{Overview of the Semantic-aware BERT model \cite{zhang2020semantics}}
\label{fig:SemBERT}
\end{figure*}

Semantic-aware BERT (SemBERT) \cite{zhang2020semantics} model has three main components, is represented in Figure \ref{fig:SemBERT}: 1) a semantic role labeler for annotating input sentences with various semantic role labels; 2) A sequence encoder that uses a pre-trained language representation model to produce a contextual representation of the input raw text and mapped semantic labels in parallel to create a semantic representation; 3) A semantic integration component that combines the text representation with the explicit contextual semantic embedding to produce an integrated representation for downstream tasks.

\textbf{\textit{Semantic Role Labeling}} is the process of assigning semantic roles to input sentences with varying semantic sequences using a pre-trained semantic labeler. There will be many predicate-argument structures for a particular sentence, such as the one in the figure \ref{fig:SemBERT}.

 \textbf{\textit{Encoding}} is a task of learning the contextual representation through the BERT model and semantics representation through the input's semantic role labels, including two parallel processes.
 \begin{itemize}
     \item Contextual Embedding: The BERT model learns the contextual representation of the input text at the subword level, then groups the subword representation word by word and uses a convolutional neural network (CNN) with max pooling to obtain the contextual representation at the word level.
     \item Semantic Embedding: These label sequences obtained from the semantic role labeler are mapped through the lookup table to form vectors and fed to the BiGRU layer to obtain a representation in the latent space, then concatenate and feed them into a fully connected class to get a semantic representation.
 \end{itemize}

 \textbf{\textit{Integration}}: Contextual representation at word level and semantic representation based on semantic role labels are concatenated together, forming a common representation of downstream tasks.

\section{Experiments and Results}

\subsection{Dataset}

\textit{ 1) SRL Task:} We use the corpus of the LORELEI Vietnamese Language Pack \cite{vilorelei} for the SRL problem. This data is collected from discussion forums, message boards, documents, social networks, and online data, including monolingual and bilingual text, vocabulary, and annotations. The dataset includes 1760 sentences that have been labeled with Simple Semantic Annotation \cite{Griffitt2018SimpleSA}. 
 
\begin{figure}[!h]
\centering
\includegraphics[scale=0.98]{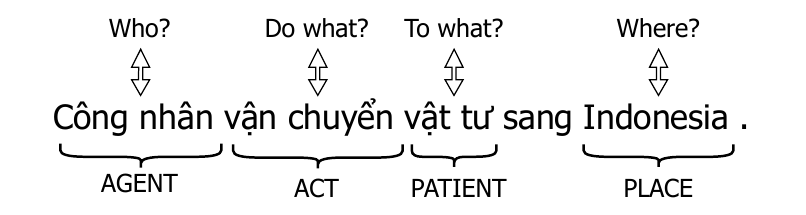}
\caption{Example of Vietnamese SRL with SSA annotation}
\label{fig:srlex}
\end{figure}
 
 \subsection{Semantic-aware BERT}

\textit{ 2) MRC Task:} UIT-ViQuAD stands for Vietnamese Question Answering Dataset, which was created specifically for the task of Vietnamese machine reading comprehension based on passages extracted from Vietnamese Wikipedia articles. Initially, UIT-ViQuAD1.0  \cite{viquad1} consisted of only answerable questions. To increase machine learning, the UIT-ViQuAD2.0  \cite{viquad2} combined 23K questions in UIT-ViQuAD1.0 with more than 12K unanswered questions. 

We separated the dataset for the MRC task into a training set, a development set, and a test set; the number of each dataset is detailed in the table \ref{tab:viquad}.

 \begin{table}[!ht]
    \centering
    \caption{Overview Statistics Of The MRC Dataset.}
    \begin{tabular}{|l|r|r|r|r|}
    \hline
    \multicolumn{1}{|c|}{\textbf{}}                                           & \textbf{Train} & \textbf{Dev} & \textbf{Test} & \textbf{All} \\ \hline
    Number of articles                                                        & 111            & 14           & 13            & 138          \\ \hline
    Number of passages                                                        & 3,038           & 637          & 426           & 4,101        \\ \hline
    Number of questions                                                       & 21,234         & 3,100        & 4,123         & 28,457       \\ \hline
    \begin{tabular}[c]{@{}l@{}}Number of unanswerable\\ questions\end{tabular} & 6,890           & 1,359         & 968           & 9,217        \\ \hline
    \end{tabular}
    \label{tab:viquad}
    \end{table}

\subsection{Evaluation Metrics}

To evaluate the performance of the model in the Vietnamese Machine Reading Comprehension task, we use two measures.

\textit{\textbf{Exact Match (EM)}} is the number of precise answers, giving a score of 1 when the prediction and the true answer are the same and 0 otherwise. When evaluating against a negative question, if the system predicts any textual span as an answer, it automatically obtains a zero score for that question.

\textit{\textbf{F1-score}} estimated over the individual tokens in the predicted answer against those in the gold standard answers is based on the number of matched tokens between the expected and gold standard answers.

\[Precision=\frac{N_{matched}}{N_{predicted}}\]

\[Recall=\frac{N_{matched}}{N_{gold\_standard}}\]

\[F1\-score=\frac{2*Precision*Recall}{Precision+Recall}\]

where $N_{matched}$ is the number of matched tokens. $N_{predicted}$ and $N_{gold\_standard}$ is the number of the predicted answer tokens and the gold standard answer tokens respectively.

\subsection{Fine-tuning SRL model}

A sentence may have more than one predicate-argument for the role labeling task. Therefore, we trained ten separate semantic role labeling models using the k-fold cross validation approach with $k = 10$ to extract various aspects of semantic information from a sentence. For each fold, we fine-tuned the XLM-RoBERTa model on 40 epochs and used AdamW \cite{DBLP:conf/iclr/LoshchilovH19} for optimization with a learning rate $1e-5$, weight decay $0.1$, and batch size $4$.

\subsection{Retro Reader}

We used monolingual and multilingual pre-training language models, such as mBERT, XLM-RoBERTa, and PhoBERT in the base version.

\begin{itemize}
    \item For the sketchy reading module, we use the following hyper-parameters learning rate $5e-6$ with the AdamW optimizer and batch size $64$, which has a cumulative gradient of $8$.
    \item For the intensive reading module, we optimally use AdamW with a learning rate of $2e-5$ and batch size $64$. The maximum length of input tokens max\_seq\_length is the maximum length of the language representation model, and the max\_query\_length query length is 64.
\end{itemize}

Finally, to integrate the two above modules, we also try to change the ratio parameter between the two models above; the thresholds are selected based on the dev set.

Additionally, we use SemBERT as a language representation model and incorporate it and its variants into Retro-sketchy Reader's reading module, fine-tuning with SemBERT's final output representation. We employ the Retro-Reader model's hyper-parameters.

\section{Results And Discussion}

Table \ref{tab:results} shows the performance evaluation of the models on the UIT-ViQuADv2.0 dataset. Accordingly, we can readily see a favorable difference between utilizing semantics and not using semantics for the Retro-Reader model. Using semantics increased the overall model evaluation score from 0.5 - 1\% on both F1\_score and EM measures. This improvement is statistically significant at $p < 0.01$ using paired t-test.

\begin{table}[!ht]
\centering
\caption{Table Of Results Of Machine Reading Comprehension Models. The Symbol \textsuperscript{\textdaggerdbl} Denotes That The Improvement Is Statistically Significant At p < 0.01 Compared With Retro-Reader’s Results Using Paired t-test.}
\label{tab:results}
\resizebox{\columnwidth}{!}{%
\begin{tabular}{|ll|c|c|}
\hline
\multicolumn{2}{|c|}{Model}                                                                                                     & EM (\%)         & F1-score (\%)   \\ \hline
\multicolumn{1}{|l|}{\multirow{3}{*}{MRC Baseline}}                                                              & mBERT        & 42.130          & 55.000          \\ \cline{2-4} 
\multicolumn{1}{|l|}{}                                                                                           & XLM-R & 45.234          & 57.872          \\ \cline{2-4} 
\multicolumn{1}{|l|}{}                                                                                           & PhoBERT      & 46.859          & 60.191          \\ \hline
\multicolumn{1}{|l|}{\multirow{3}{*}{Retro-Reader}}                                                              & mBERT        & 44.118          & 55.858          \\ \cline{2-4} 
\multicolumn{1}{|l|}{}                                                                                           & XLM-R & 48.072          & 60.521          \\ \cline{2-4} 
\multicolumn{1}{|l|}{}                                                                                           & PhoBERT      & 48.363          & 61.262          \\ \hline
\multicolumn{1}{|l|}{\multirow{3}{*}{\begin{tabular}[c]{@{}l@{}}Retro-Reader\\   + Semantic (CLS)\end{tabular}}} & mBERT       & 44.967\textsuperscript{\textdaggerdbl}          & 56.326\textsuperscript{\textdaggerdbl}          \\ \cline{2-4} 
\multicolumn{1}{|l|}{}                                                                                           & XLM-R & 48.485\textsuperscript{\textdaggerdbl}          & 61.065\textsuperscript{\textdaggerdbl}          \\ \cline{2-4} 
\multicolumn{1}{|l|}{}                                                                                           & PhoBERT     & \textbf{49.032\textsuperscript{\textdaggerdbl}} & \textbf{61.896\textsuperscript{\textdaggerdbl}} \\ \hline
\end{tabular}%
}
\end{table}

The results of the Sketchy Reading Modules of each model in the answerability question classification task are shown in Table \ref{tab:result-cls}. As we can see, applying semantic from semantic role labels to the three language models, mBERT, XLM-RoBERTa, and PhoBERT, yielded an increase of 4.341\%, 1.9\%, and 1.414\%, respectively, compared to the conventional contextual model in the task of identifying unanswerable questions. This helps to improve the total score for the Retro-Reader model on the Machine Reading Comprehension.

 \begin{table}[!ht]
\centering
\caption{The Results Of The Sketchy Reading Module's Unanswerable Question Verification Task.}
\label{tab:result-cls}
\resizebox{0.9\columnwidth}{!}{%
\begin{tabular}{|l|c|c|}
\hline
                    & Accuracy (\%) & F1-score (\%) \\ \hline \hline
BERT                & 71.258   & 48.275   \\ \hline
BERT + Semantic    & 73.490   & 55.518   \\ \hline \hline
XLMR                & 76.110   & 62.188   \\ \hline
XLMR + Semantic    & 76.985   & 66.695        \\ \hline \hline
PhoBERT             & 73.805   & 53.333   \\ \hline
PhoBERT + Semantic & 75.036   & 56.954        \\ \hline
\end{tabular}%
}
\end{table}

According to the results for the SRL task in table \ref{tab:result_srl}, we can see that the F1-score of the models is poor, which can be explained by the amount of SRL data in the LORELEI dataset we used being relatively limited and not enough to train a good labeling model. This can influence the labeling of input sentences, and inaccurate semantic labels can damage the extraction of semantic information; thus, we anticipate these improvements could be significantly better if we trained a more accurate SRL model.

\begin{table}[!ht]
\centering
\caption{The Results Of The SRL Problem After Fine-Tuning XLM-R, On The LORELEI Dataset.}
\label{tab:result_srl}
\resizebox{0.85\columnwidth}{!}{%
\begin{tabular}{|c|c|c|c|}
\hline
\multicolumn{1}{|l|}{}  & Precision (\%)  & Recall (\%)  & F1-score (\%)  \\ \hline
Fold 1                  &  35.261        &  32.568        & 33.871       \\ \hline
Fold 2                  &  34.700        &   30.375        & 32.393      \\ \hline
Fold 3                  &  38.776        &   32.095         & 35.120      \\ \hline
Fold 4                  &  41.958       &   29.557      & 34.682       \\ \hline
Fold 5                  &  38.821        &  31.728      & 34.917      \\ \hline
Fold 6                  &  40.144       &  34.901         & 34.900      \\ \hline
Fold 7                  &  36.719       &   32.192      & 34.307      \\ \hline
Fold 8                  &  39.535        &   24.286      & 36.724       \\ \hline
Fold 9                  &   39.241      &   32.518      & 35.564      \\ \hline
Fold 10                 &  39.448        &   34.014      & 36.530       \\ \hline
Average                 &  38.460           &  31.423       & 34.901      \\ \hline
\end{tabular}%
}
\end{table}

Overall, the Retro-Reader model improvement ranges from 0.5\% to 1\%, which is insignificant for the Vietnamese MRC task, but our experiment has partly shown the potential of semantic extraction from semantic role labeling in Vietnamese. The positive results of this experiment will serve as the basis for future research into the usage of SRL for other NLU tasks in Vietnamese.

\section{Conclusion}

\subsection{Summary}
In this study, we mainly compare the effect of semantics on the classification of answerability questions for the Vietnamese MRC problem by integrating semantic information into the encoder of the sketchy reading module in Retro-Reader. The results show that the base version mBERT, XLM-R, and PhoBERT models with semantic information perform better than conventional contextual representation models. We have described the implementation approach and experimental results for the MRC problem. In particular, the results improved by 1.4 - 4.3\% for the Retro-Reader model's Sketchy Reading Module in the sub-task of classifying the question's answerability, helping the model's overall results improve by 0.5 - 1\%.  This is one of the earliest research utilizing semantic information from semantic labels for language representation in Vietnamese. As a result of the acquired experimental results, it is possible to realize the potential of semantic extraction from SRL for various NLU tasks in Vietnamese.

However, during the implementation of this experiment, we encountered some difficulties in training pre-trained SRL models because of the limitation of the corpus size. This has affected the annotation of semantic role labels for the input sentences. And based on the encouraging results of this study, we want to partially encourage further research on the SRL task for Vietnamese, as there are currently few research works and corpora for this task in Vietnamese.

\subsection{Future Work}

We identify several directions for further research and development based on the results and study limitations, including:

\begin{itemize}
    \item Due to time and resource limitations, we have only tested the pre-trained language model's base version and a small number of variants. In the future, we will try more semantic-aware BERT model variants with a larger version.
    \item In this study, we have only focused on improving the answerability classification task, but in the future, we will be testing semantic integration into Retro-Reader’s Intensive Reading Module.
    \item We believe that if there is enough data to train an SRL model, the input sentence labeling will be enhanced, allowing the semantic integration model to perform better. As a result, in the future, we will create a semantic role labeling dataset to train an SRL model that is more accurate than the existing model.
\end{itemize}

\section*{Acknowledgement}
This research is funded by University of Information Technology-Vietnam National University HoChiMinh City under grant number D1-2022-47.

\bibliographystyle{IEEEtran}
\bibliography{main.bib}

\end{document}